\documentclass[journal]{IEEEtai}
\usepackage{tikz,graphics,color,fullpage,float,epsf,caption,subcaption}
\usepackage{cite}
\usepackage{graphicx}
\usepackage{amsmath}
\usepackage{amssymb}
\usepackage{booktabs}
\usepackage{algorithm}
\usepackage{algorithmic}
\usepackage{multirow}
\usepackage[colorlinks,urlcolor=blue,linkcolor=blue,citecolor=blue]{hyperref}
\newcommand{\etal}{\textit{et al}. }

\begin{document}

\title{Balancing Uncertainty and Diversity of Samples: Leveraging Diversity of Least, High Confidence Samples for Effective Active Learning}

\author{
Vipul Arya, S.H. Shabbeer Basha \IEEEmembership{Member,~IEEE}, Srikrishna U N, Sunainha Vijay, and Snehasis Mukherjee \IEEEmembership{Senior Member,~IEEE}

\thanks{V. Arya, Srikrishna, and S. Vijay are with the School of Computer Science and Engineering, RV University, Bengaluru, Karnataka-560059, India (e-mail: krishnamvipul@gmail.com,
srikrishnaun.btech22@rvu.edu.in,  sunainhav.btech22@rvu.edu.in). }
\thanks{S.H.S. Basha is with the School of Engineering \& Technology, Vidyashilp University, Bangalore, India (e-mail: shabbeer.basha@vidyashilp.edu.in). }
\thanks{S. Mukherjee is with Shiv Nadar Institution of Eminence, Delhi NCR (e-mail: snehasis.mukherjee@snu.edu.in). }
}

\maketitle

\begin{abstract}
Deep learning models, including Convolutional Neural Networks (CNNs) and Vision Transformers (ViTs), have achieved state-of-the-art performance on various computer vision tasks such as object classification, detection, segmentation, generation, and many more. However, these models are data-hungry as they require more training data to learn millions or billions of parameters. Especially for supervised learning tasks, curating a large number of labeled samples for model training is an expensive and time-consuming task. Active Learning (AL) has been used to address this problem for many years. Existing active learning methods aim to choose samples for annotation from a pool of unlabeled samples that are either diverse or uncertain. Choosing such samples may hinder the model's performance, as we pool based on one dimension, i.e., either diversity or uncertainty. In this paper, we propose four novel hybrid sampling methods for pooling both easy and hard samples, which are also diverse. To verify the efficacy of the proposed methods, extensive experiments are conducted using high and low-confidence samples separately. We observe from our experiments that the proposed hybrid sampling method, Least Confident and Diverse (LCD), consistently performs better compared to state-of-the-art methods. It is observed that selecting uncertain and diverse instances helps the model learn more distinct features. The code related to this study will be available at \url{https://github.com/vipularya135/LCD}.
\end{abstract}

\begin{figure}
\centerline{\includegraphics[width=0.5\textwidth]{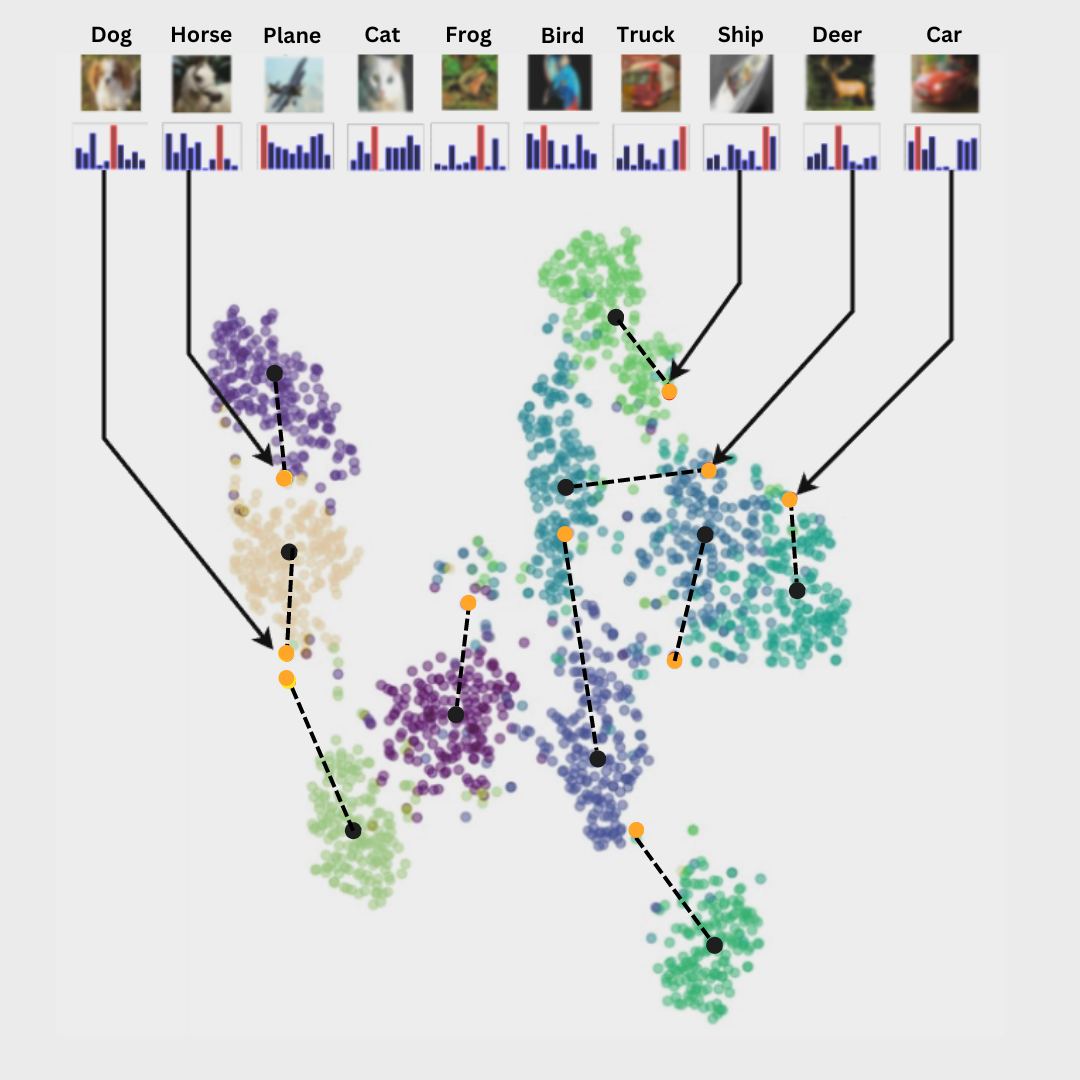}}
\caption{Overview of the proposed LCD Active Learning framework. First, the hard samples with the least confidence scores (which are highlighted in red in the histogram) are chosen from the unlabeled pool. Then, the top $B$ diverse samples, which are far away from the cluster center, are selected from these least confidence samples. For instance, we prefer to consider a sample that belongs to the horse category and is close to the decision boundary (highlighted in orange), which is also away from the cluster center, rather than considering samples that are closer to the cluster center for the horse category.}
\label{figure1}
\end{figure}

\begin{IEEEImpStatement}
Supervised learning models require a huge amount of annotated data for training, which is difficult in many applications. Active learning methods enable models to improve their ability to learn with fewer annotated data. A typical active learning approach starts with a few annotated data to train a sophisticated model. The small set of data is insufficient to train a large model. We predict the levels of the rest of the unlabeled samples using the trained model. The goal of any active learning approach is to iteratively select the next batch of samples to fine-tune the trained model. Existing active learning approaches try with different combinations of high- and low-confidence samples. This study focuses on popular hybrid sampling strategies, including diversity (i.e., different from the same kind of images). Experimental results on benchmark datasets demonstrate that the proposed diverse sampling methods can lead to better performance.
\end{IEEEImpStatement}


\section{Introduction}
\label{sec:intro}
Convolutional Neural Networks (CNN) and Vision Transformers (ViT) have become popular deep neural network architectures in several domains, especially computer vision. This is because CNNs and ViTs naturally understand how to capture spatial features and dependencies in local and global contexts, respectively. This makes them better than other suggested architectures for the same tasks. Many attempts have been made in the literature to improve the performance of neural networks by proposing different variants of CNNs \cite{simonyan2015very,he2016deep,huang2017densely} and transformers \cite{liu2021swin,dosovitskiy2020image}. However, the performance of a deep learning model does not solely depend on the network architecture but also on the training dataset's size and quality. According to the literature, a diverse and comprehensive dataset leads to descent performance on respective tasks \cite{deng2009imagenet,lin2014microsoft}. While solving a supervised problem such as object recognition, object detection, the model requires a huge amount of labeled data for training. However, it is time-consuming, labor-intensive, and not cost-efficient to annotate the entire unlabeled data. These limitations raise an intriguing question: \textit{How to smartly choose the right samples to annotate so that a descent performance is obtained given a labeling budget.}

In the past, semi-supervised and unsupervised learning models have been applied to address the challenges of annotating large unlabeled dataset \cite{ebert2012semi,gao2020consistency,le2016budgeted,rebuffi2020semi}. The uniqueness of active learning lies in the selection of the most important data points based on some criteria, which can effectively increase the model's accuracy while reducing the number of labeled samples for model training. Gathering an initial set of labeled data for model training is the first step in active learning. The model trained on these initial samples is considered as a candidate model to sample the next set of images from the unlabeled dataset. Then, we select the most useful and informative data points using effective data selection strategies. There have been different ideas in the past about how to choose samples based on diversity \cite{agarwal2020contextual,wang2017incorporating}, uncertainty \cite{li2006confidence,raj2022convergence,wang2017uncertainty}, entropy \cite{siddiqui2020viewal}, and other factors. The selected data points are the samples for which the model is uncertain. The labels for these data points are obtained manually. These samples are combined with the initial samples, using which the model is retrained. We repeat this process until we exhaust the labeling budget. Figure \ref{figure1} provides an overview of the proposed active learning process.

The main challenge in any active learning paradigm is to choose the optimal values for dynamic parameters such as uncertainty thresholds, diversity, and various other parameters that affect the active learning process. Setting these parameters at the beginning of the learning process can be tricky. Another challenge in active learning is the biased nature of datasets. During the initial model training, obtaining a balanced dataset is difficult. While CNNs learn from data in batches, traditional active learning methods select individual data points that the model is uncertain about and learn them, which often interrupts the model's efficiency \cite{sener2018active}.  

Several attempts have been made to develop robust sampling techniques for active learning \cite{agarwal2020contextual,raj2022convergence,wu2022entropy}. Uncertainity based active learning methods often lack diversity \cite{raj2022convergence}. For example, marginal confidence sampling \cite{nguyen2022measure}, least confidence sampling \cite{monarch2021human} methods aim at selecting the ambiguous and least confident samples. This observation motivated us to develop new methods that effectively handle the trade-off between uncertainty and diversity to choose better representative samples. A similar attempt has been made in the literature by Yang \etal \cite{yang2015multi} to select the uncertain samples, which are diverse as well. However, the process used select diverse cum uncertain samples is expensive, especially working with higher dimensional data. Moreover, the efficacy of this hybrid active learning method is verified on traditional machine learning models like SVM.  

We propose simple but effective sampling methods for deep active learning. We experiment with sampling methods that are aimed at i) diversity and uncertainty. ii) diversity and certainty iii) uncertainty and certainty, and iv) uncertainty, certainty, and diversity. This study investigates the above four distinct methods for expanding the training dataset. Initially, a model is trained on a fixed $4\%$ of samples, and subsequently, the same model undergoes retraining using the samples selected by one of the above four hybrid sampling methods. In summary, this paper's contributions are as follows:
\begin{itemize}
\item Designed four novel and hybrid sampling methods for active learning.
\item Conducted extensive experiments with various CNN and transformer models to verify the efficacy of the proposed sampling methods.
\item Provided comprehensive analysis and insights on the comparative performance of the sampling methods.
\end{itemize}

\section{Related works}
Active learning (AL) aims to improve model performance with fewer labeled samples, making it valuable when labeling is costly or time‐intensive. It has been widely applied to various tasks such as object recognition \cite{long2015fully,tylevcek2013spatial}, object detection \cite{liu2018d3r,wu2022entropy}, and regression \cite{jose2024regression,kumar2020active} problems. The primary objective of active learning is to make machine learning models perform better with fewer training samples. This selective learning approach is useful in situations where unlabeled samples are available, but obtaining labels is highly expensive and time-consuming. Early works in active learning, which include queries and concept learning by Angluin \etal \cite{angluin1988queries} and Cohn \etal \cite{cohn1994improving}, laid the foundation by exploring different query strategies and problem settings.

The most common strategies in active learning are pool-based methods that pick up the most useful examples from a large set of unlabeled data. The most popular pool-based active learning is uncertainty sampling \cite{lewis1994sequential,tong2001support}, wherein the learner queries the instances about which they are least confident. Lewis \etal \cite{lewis1994sequential} validated the effectiveness of uncertainty sampling in the context of textual data. Their study suggests that a model trained with fewer samples pooled by an active learning strategy can achieve similar or better results compared to training with an entire dataset. Various machine learning models, including logistic regression and support vector machines (SVMs), have extensively adapted this technique. For example, Tong \etal \cite{tong2001support} conducted uncertainty sampling on SVMs, highlighting its effectiveness in reducing the labeling effort while retaining the model performance. Settles \etal \cite{settles2008analysis} found that pool-based active learning can greatly reduce the amount of labeling needed without compromising performance. Tong \etal \cite{tong2001support} proved the effectiveness of the active learning method in image retrieval.

Another popular active learning technique is the Query-by-Committee (QBC), in which a committee of models representing different hypotheses based on the labeled data is used to select the most informative samples \cite{seung1992query}. The learner then asks for examples about which models in the committee disagree most, focusing on the most informative questions. McCallum \etal \cite{mccallum1998employing} extended the QBC method \cite{seung1992query} by using a pool of unlabeled documents to estimate document density, thus selecting the documents from denser regions for labeling. They further integrate active learning with expectation maximization (EM) to infer class labels for unlabeled documents. Their experiments show significant reductions in the number of labeled examples required to achieve high accuracy, compared to using the traditional QBC or EM alone. This approach has been proven to be effective in various applications, including text classification and data extraction.

A few studies have explored hybrid strategies integrating multiple-question strategies to leverage their respective strengths. Hoi \etal \cite{hoi2006batch} proposed a hybrid active learning method that combines uncertainty sampling with other techniques to improve learning efficiency. This method was evaluated on tasks such as image and video classification \cite{sun2019active}, where different strategies can work together to achieve better results. Furthermore, researchers have developed advanced methods such as deep active learning, which use deep neural networks as base models. Gal \etal \cite{gal2017deep} introduced a Bayesian approach to active learning with deep learning models, using dropout as a way to estimate uncertainty.

In a conventional active learning setting, the sampling algorithm acquires a label for one sample at a time, which is infeasible while training deep learning models. In batch-mode active learning (AL) approaches \cite{guo2007discriminative}, numerous instances are selected for labeling at a time rather than just a single instance. It is a faster approach with less computational overhead. Batch mode AL strategies use clustering or ranking methods to select a diverse and fluorescent batch of instances. Guo \etal \cite{guo2007discriminative} demonstrated the strength of this idea in decreasing runtime cost by lowering convergence time to the desired expected precision. Core set selection \cite{sener2018active} is an active learning method that selects a subset that yields the best model performance. They have demonstrated the efficacy of active learning with deep learning models by leveraging core sets to obtain a high level of training efficiency in neural networks, but with relatively little compromise on accuracy.

Adversarial Active Learning trains a generative 
adversarial network (GAN), where the generator generates hard examples that the current model fails to predict. The hard samples are added to the pool, which is later used for querying and labeling. Adversarial active learning \cite{zhu2017generative} achieved high accuracy with fewer labeled samples compared to other pool based active learning methods.

Semi-supervised active learning \cite{huang2021semi} is an integration of semi-supervised algorithms and active learning methods. It uses both labeled and unlabeled data to make training more efficient. The active learner queries the most informative samples, and this works with a semi-supervised learner, which serves as a way for maximizing all that unlabeled data. This hybrid method allows for dramatically lower labeling costs while preserving model performance. Recent research explores ways to make reinforcement learning more efficient by combining it with active learning methods \cite{dodds2024sample}.


Active learning for regression applications \cite{jose2024regression} has focused a lot of emphasis on reducing training set size without sacrificing performance. Even though classification has been the subject of extensive research, regression poses unique challenges due to its intricacy. Jose \etal \cite{jose2024regression} proposed an active learning strategy based on regression trees to enhance existing methods. By combining model-free techniques and maximizing information utilization, this strategy achieves exceptional results across small and large training sets, demonstrating improved performance consistency over several experiments.

More recently, efforts to improve labeling accuracy in active learning for data clustering have concentrated on selecting and splitting data blocks optimally \cite{wang2020active}. The ensemble-based active learning technique breaks up data into blocks and carefully searches labels for important examples. Identifying the most effective strategies for blocking and selecting data presents significant challenges.

In summary, active learning has evolved drastically over the past few years, with numerous strategies and frameworks advanced to cope with distinctive factors of the learning technique. From uncertainty sampling and QBC to hybrid approaches and deep learning, the sphere has made substantial strides in lowering labeling costs and enhancing version performance across numerous packages. Because the availability of large unlabeled datasets continues to grow, active gaining knowledge of duties remains a crucial area of study, with ongoing developments geared toward improving its efficiency and effectiveness in sensible device gaining knowledge of duties. Despite the recent efforts to sample unlabeled data in an active learning paradigm, a proper investigation of the effects of different sampling methods is missing in the literature. This study conducts a thorough analysis on the performances of the different hybrid sampling methods.

\section{Methodology}
We first discuss the mathematical formulation of this study, followed by the different sampling methods.

\begin{algorithm}
\caption{Procedure for the proposed hybrid Active Learning framework}
\begin{algorithmic}
\label{al_alg}
\STATE \textbf{Given:} Labeled set $\mathcal{L}$, unlabeled pool $\mathcal{U}$, query strategies $\phi_{\text{p}}(x)$, $\phi_{\text{q}}(I)$, query batch size $B$
\REPEAT 
    \STATE // Learn a model using the current $\mathcal{L}$
    \STATE $\theta = \text{train}(\mathcal{L})$
    \WHILE {$b$ = 1 \textbf{to} $B$ }
        \STATE // Query the instances from the pool based on first criterion
        \STATE $I = \arg \max_{x \in \mathcal{U}} \phi_{\text{p}}(x)$
    \STATE // Apply second criterion function
    \STATE $S = \arg \max_{I \in \mathcal{U}} \phi_{\text{q}}(I)$
    \STATE // $S$ represents set of samples selected in one iteration  
    \STATE // Move the labeled query from $\mathcal{U}$ to $\mathcal{L}$
    \STATE $\mathcal{L} = \mathcal{L} \cup \{ S, \text{label}(S) \}$
    \STATE $\mathcal{U} = \mathcal{U} - S$
    \ENDWHILE
\UNTIL{some stopping criterion}
\end{algorithmic}
\end{algorithm}

\subsection{Background}
Let us consider $\mathcal{U}$ denotes the pool of unlabeled samples, and $\mathcal{L}$ is the initial set of labeled samples used for model training. Here, $\mathcal{L}\subset \mathcal{U}$, which has initial samples chosen randomly and annotated for model training. The model trained on initially pooled samples is considered to be a candidate model using which the next set of samples can be selected. The remaining number of unlabeled samples is $\mathcal{U}$-$\mathcal{L}$. Now, to select the optimal set of samples from the remaining unlabeled dataset, a criterion function (sampling method) $\phi(.)$ is used. In each iteration of the proposed hybrid sampling methods, $\mathcal{B}$ number of samples represented by $\mathcal{S}$ are selected for updating the model. Throughout our experiments, $B$ is considered to be 2500. Initial training samples $\mathcal{L}$ are 2000 (which remain the same across all the experiments), and 2500 samples are added to $\mathcal{L}$ in each active learning iteration. 

Algorithm \ref{al_alg} describes the proposed novel hybrid active learning algorithm. The two criterion functions $\phi_{\text{p}}(x)$ and $\phi_{\text{q}}(I)$ are applied during the first and second stages as follows in the different sampling methods.

We apply different sampling strategies (low and high confidence) and their combinations in the proposed active learning paradigm. We employ diverse sampling following \cite{dietterich2000ensemble}, where we cluster the features and rank them according to the distance from the cluster centre, as several previous studies rely on this cluster-based diversity measure \cite{chavhan2023quality}.

\subsection{High Confidence and Diverse Sampling (HCD)}
HCD selects high‐confidence samples that are also diverse:
\begin{enumerate}
\item From the remaining unlabeled pool, choose samples with maximum predicted class probability (Eq. 1). This means, we select the samples $x^*$ as follows:
\begin{equation}
\label{eq1}
x^*=\arg\max_x(P_{\theta}(\hat{y}|x))
\end{equation}
where $x$ are the remaining samples, and $\hat{y}$ represents the highest posterior probability predicted by model $\theta$.
\item Next, we consider diverse samples from these selected ones, based on their distance from the cluster center. KMeans clustering \cite{macqueen1967some} is used to form clusters, and Euclidean distance is used to calculate the distance from the cluster center. The finally selected sample $x^{**}$ is obtained as follows:
\begin{equation}
\label{eqd}
x^{**}=\arg\max_{d_{x^*,x_c}}x^*
\end{equation}
where $x_c$ is the cluster center for the sample $x^*$ selected by equation (\ref{eq1}), and $d_{x^*,x_c}$ is the Euclidean distance between $x^*$ and $x_c$.
\end{enumerate}

The HCD sampling method selects the instances at which the model is confident, and they are also diverse compared to the current training set. This type of hybrid sampling allows diversity to be added to the training set. In each iteration, 2.5K samples are added to current training set for model improvement.

\subsection{Low Confidence and High Confidence (LCHC) Sampling}
This sampling method selects both low-confidence and high-confidence instances for labeling:
\begin{enumerate}
\item First, the 1.25k samples from the remaining unlabeled set are identified with the least confidence scores predicted by the model, indicating high uncertainty. This means, we select the samples $x^*$ as follows:
\begin{equation}
\label{eq2}
x^*=\arg\max_x(1-P_{\theta}(\hat{y}|x))
\end{equation}
where $x$ are the remaining samples, and $\hat{y}$ represents the highest posterior probability predicted by model $\theta$.
\item Next, we find another 1.25k data samples with the highest confidence scores following (\ref{eq1}).
\end{enumerate}
In each iteration, 2.5K samples are selected, annotated, and used for model improvement.

\begin{table}[]
\centering
\caption{Dataset details: $|\mathcal{U}|$, $|\mathcal{L}|$, and $|\mathcal{B}|$ represent the cardinality of the pool of unlabeled samples, the initial training set (4\% of the data), and the batch of samples selected for labeling in each iteration (5\% of the data), respectively.}
\begin{tabular}{|l|l|l|l|}
\hline
Dataset & $|\mathcal{U}|$ & $|\mathcal{L}|$ (4\%) & $|\mathcal{B}|$ (5\%)  \\ \hline
CIFAR-10 \cite{krizhevsky2009learning} & 50000 &  2000 & 2500  \\ \hline
CIFAR-100 \cite{krizhevsky2009learning} & 50000 & 2000 & 2500  \\ \hline
SVHN \cite{netzer2011reading} & 73257 & 2930 & 3663  \\ \hline
\end{tabular}
\label{table_dataset_details}
\end{table}

\subsection{Dual-Strategy Active Learning (DSAL)}
This sampling method selects diverse easy (with high confidence) and hard (uncertain) instances from the remaining unlabeled data as follows:
\begin{enumerate}
\item First, top uncertain (least confidence) samples are selected following equation (\ref{eq2}). From this, the samples that are far away from the cluster center (diverse) are considered. 
\item Next, from the top easy samples following equation (\ref{eq1}), the diverse samples are selected.
\end{enumerate}

Note that KMeans clustering \cite{macqueen1967some} is used to form clusters (the number of classes is considered to be the $K$ value), and Euclidean distance is used to calculate the distance from the cluster center.  Throughout the DSAL method experiments, we consider i) 1.25k samples as hard (uncertain) and diverse, ii) 1.25k samples as easy and diverse, and ultimately select 2.5k samples for labeling and model training. We add these 2.5k samples to the initial 2k samples in the first iteration and then add them to the previously aggregated samples for each subsequent iteration.

\subsection{Low Confidence and Diverse (LCD) Sampling}
Similar to HCD, the LCD sampling method selects the instances for annotation iteratively as follows:

\begin{enumerate}
\item Identify least‐confident predictions (Eq. \ref{eq2}).
\item Next, from these selected samples, retain those farthest from their cluster centers (Eq. \ref{eqd}). This yields uncertain, informative, and varied samples, improving generalization.

\end{enumerate}

The rationale behind selecting the diverse sample is that, the diverse samples are, being away from their cluster centers, much different from other samples, which makes them carry more information to the model for training.

The LCD sampling method selects the instances at which the model is very uncertain, and they are also diverse compared to the current training set. This kind of hybrid sampling enables the addition of both uncertain and diverse instances to the training set. We select 2.5K samples for labeling and model training in each iteration. \\

For all the above four sampling methods, both high‐confidence and low‐confidence strategies, we first obtain the confidence scores for all unlabeled samples. The samples are then ranked based on their confidence scores, where the top-$\mathcal{L}$ ranked samples correspond to high confident predictions, and the bottom-$\mathcal{L}$ ranked samples correspond to the least confident (most uncertain) predictions. From which top $4\%$ diverse samples are selected in each active learning iteration.

\section{Experimental Results}
We first illustrate the experimental setup, followed by a detailed discussion of the results.

\subsection{Experimental setup}
To establish the efficacy of the proposed hybrid sampling methods, extensive experiments are performed on popular deep learning models such as: i) VGG-16 \cite{simonyan2015very}, ResNet-18 \cite{he2016deep}, ResNet-50 \cite{he2016deep}, MobileNet \cite{howard2017mobilenets}, DenseNet-121 \cite{huang2017densely}, Swin Transformer \cite{liu2021swin}, and ViT-Small \cite{dosovitskiy2020image} transformer models using CIFAR-10 . ii) ResNet-18 \cite{he2016deep}, ResNet-50 \cite{he2016deep}, DenseNet-121 \cite{huang2017densely}, and Swin transformer \cite{liu2021swin} models using CIFAR-100. and iii) VGG-16 \cite{simonyan2015very}, ResNet-18 \cite{he2016deep}, ResNet-50 \cite{he2016deep}, and ResNet-56 \cite{he2016deep} models using SVHN dataset. The details, such as unlabeled pool size, initial training set size, and batch of samples selected in each active learning iteration for CIFAR-10 \cite{krizhevsky2009learning}, CIFAR-100 \cite{krizhevsky2009learning}, and SVHN \cite{netzer2011reading}, are given in Table \ref{table_dataset_details}.

\begin{table}[]
\centering
\caption{Performance comparison of the proposed hybrid active learning methods compared against the state-of-the-art on CIFAR-10. Bold denotes the active learning method securing the highest accuracy for the corresponding deep network.}
\begin{tabular}{|l|p{0.3\linewidth}|r|}
\hline
\textbf{Models} & \textbf{Methods} & \textbf{Test Accuracy} \\
\hline
\multirow{8}{*}{\textbf{VGG-16 \cite{simonyan2015very}}} 
  & Core set   \cite{sener2018active}   & 90\%    \\
  & VAAL                 \cite{sinha2019variational}       & 90.16\% \\
  & Influence AL     \cite{liu2021influence}    & 90\%    \\
  & HCD                & 93.02\% \\
  & LCHC                     & 93.64\% \\
  & DSAL          & 93.83\% \\
  & LCD                & \textbf{93.87\%} \\
\hline
\multirow{6}{*}{\textbf{ResNet-18 \cite{he2016deep}}} 
  & DEAL \cite{hemmer2021deal}            & 89\%    \\
  & VAAL \cite{sinha2019variational}            & 81\%    \\
  & Random AL \cite{ji2023randomness}                & 91\% \\
  & CoreGCN \cite{caramalau2021sequential}                    & 90.7\% \\
  & CB AL \cite{bengar2022class}                & 91\% \\
  & HCD                & 94.55\% \\
  & LCHC                     & 94.83\% \\
  & DSAL           & 94.91\% \\
  & LCD                & \textbf{95.00\%} \\
\hline
\multirow{5}{*}{\textbf{ResNet-50 \cite{he2016deep}}} 
  & DRL \cite{sun2019active}         & 90\%    \\
   & EOAL  \cite{zong2026revisiting}    & \textbf{95.94\%}    \\
  & HCD                & 95.27\% \\
  & LCHC                     & 95.48\% \\
  & DSAL           & 94.91\% \\
  & LCD                & \textbf{95.00\%} \\
\hline
\multirow{7}{*}{\textbf{ResNet-56 \cite{he2016deep}}} 
  & AS    \cite{zhang2022improving}      & 88.86\% \\
  & CB AL \cite{bengar2022class} & 91 \\
  & HCD                & 95.11\%    \\
  & LCHC                     & 95.25\% \\
  & DSAL          & 95.09\%  \\
  & LCD                & \textbf{95.26\%}    \\
\hline
\multirow{7}{*}{\textbf{MobileNet \cite{howard2017mobilenets}}} 
  & FPGA               \cite{bouguezzi2021efficient}         & 86\%    \\
  & AOT              \cite{jindal2024army}           & 80\%    \\
  & PA-CL               \cite{chen2024manipulating}        & 82\%    \\
  & HCD                & 90.16\% \\
  & LCHC                     & 91.14\% \\
  & DSAL           & 90.85\% \\
  & LCD                & \textbf{91.39\%} \\
\hline
\multirow{6}{*}{\textbf{DenseNet-121 \cite{huang2017densely}}} 
  & Triplet-AL           \cite{sundriyal2021semi}       & 75\%    \\
  & EAL         \cite{jindal2024army}        & 75.3\%  \\
  & HCD                & 94.99\% \\
  & LCHC                     & 95.50\% \\
  & DSAL           & 95.30\%  \\
  & LCD                & \textbf{95.35\%} \\
\hline
\multirow{6}{*}{\textbf{Swin \cite{liu2021swin}}} 
  &  MESA            \cite{pan2022mesa}         & 81.3\%    \\
  & GradGAT            \cite{zhang2022nested}         & 78\%    \\
  & EswinT           \cite{Yao2024data}           & 81\%    \\
  & HCD                & 82.95\% \\
  & LCHC                     & 83.89\% \\
  & DSAL           & 83.61\% \\
  & LCD                & \textbf{85.40\%} \\
\hline
\multirow{6}{*}{\textbf{ViT-Small \cite{dosovitskiy2020image}}} 
  & SeqPool             \cite{hassani2022escaping}        & 78.4\%  \\
  & HCD                & 81.35\% \\
  & LCHC                     & 82.10\%\\
  & DSAL           & 82.33\% \\
  & LCD                & \textbf{83.05\%}  \\
\hline
\end{tabular}
\label{table_cifar-10_results}
\end{table}

All the above active learning experiments are conducted using model-dataset combinations with the following configuration of hyper-parameters. The CNN and transformer models are trained for 50 epochs, considering the initial value of the learning rate as 0.01. The learning rate decays by a factor of 0.1 for every 10 epochs. 

\subsection{Results and Discussions}
The results of the proposed active learning method are shown in tables \ref{table_cifar-10_results}, \ref{table_cifar-100_results}, \ref{table_SVHN_results}, \ref{table_tiny_results} and \ref{table_pascal_results}. These tables compare the results of the proposed method to the state-of-the-art active learning methods on the CIFAR-10, CIFAR-100, SVHN, Tiny ImageNet and Pascal VOC 2012 datasets, respectively. From these tables, it is evident that the proposed hybrid active learning method LCD outperforms all the other competing sampling methods, including the recent state-of-the-art.

Table \ref{table_cifar-10_results} presents a comparison of various active learning results on CIFAR-10. The proposed hybrid active learning variants, particularly LCD, achieve improved results compared to several benchmarks such as VAAL \cite{sinha2019variational}, influence selection \cite{liu2021influence}, and others. For instance, on the VGG-16 model, the proposed LCD sampling method achieves a test accuracy of 93.87\%, outperforming the baseline methods. Overall, the best test accuracy of 95.35\% is obtained by employing LCD for training DenseNet-121.

Table \ref{table_cifar-100_results} summarizes the results on the CIFAR-100 dataset. Consistently, the proposed sampling methods show improved results on CIFAR-100 compared to the recent state-of-the-art. For example, on the DenseNet-121 model, the proposed LCD method achieves the best test accuracy of 78.49\%, surpassing benchmark methods such as BNN \cite{rakesh2021efficacy} and Lattice-Cross Fusion \cite{dealmeida2021turning} by a significant margin.

Active learning results for the SVHN dataset are presented in Table \ref{table_SVHN_results}. The proposed LCD sampling method has obtained better results than all the competing methods and other sampling schemes. Tables \ref{table_tiny_results} and \ref{table_pascal_results} show results on Tiny ImageNet and PASCAL VOC 2012. LCD achieves the best accuracy across all models, e.g., 66.57\% with ResNet-50 on Tiny ImageNet and 84.44\% with VGG-16 on PASCAL VOC, outperforming existing results. \\

However, compared to the deep reinforcement learning (DRL) method \cite{sun2019active}, the LCD method achieved comparable results. While our experiments are on image classification and object detection benchmarks (CIFAR, SVHN, TinyImageNet, PASCAL VOC), the method is architecture-agnostic and can be adapted to text, sensor, time-series, and video domains.

\subsubsection{Potential Domain-Specific Applications and Impact}
Although our evaluation has focused on standard vision benchmarks, the proposed hybrid active learning strategies are broadly applicable across various domains where labeling is expensive. In medical imaging, uncertain-diverse selection can surface rare pathological cases (e.g., early tumor formations) for expert review, enabling efficient model improvement. In autonomous driving, our approach can prioritize atypical traffic situations and challenging visual conditions, enhancing safety-critical perception models. Furthermore, remote sensing can benefit through targeted labeling of distinct environmental changes or rare geographical features, while industrial visual inspection can identify diverse defect types in manufacturing pipelines. The flexibility of our strategy allows it to be integrated into any task where iterative annotation and model retraining are viable, enabling faster deployment of robust AI systems in domain-specific workflows.

Although this work primarily aims to enhance the effectiveness of hybrid sampling methods through the integration of uncertainty and diversity, it is important to recognize that active learning frameworks may be vulnerable to adversarial poisoning. Such attacks involve injecting harmful samples into the dataset to impair the model’s accuracy or induce biased results. To improve resilience against these threats, techniques like anomaly detection, identification and removal of outliers, and adversarial training can be employed. Moving forward, we plan to explore the development of comprehensive approaches that unite robust sampling strategies with advanced protective measures, ensuring both high performance and security in sensitive applications.

\begin{table}[]
\centering
\caption{Performance of the proposed hybrid sampling methods compared against the state-of-the-art active learning methods applied to CIFAR-100. Bold denotes the sampling scheme securing the highest accuracy for the corresponding deep network.}
\begin{tabular}{|l|p{0.3\linewidth}|r|}
\hline
\textbf{Models} & \textbf{Methods} & \textbf{Test Accuracy} \\
\hline
\multirow{9}{*}{\textbf{ResNet-18 \cite{he2016deep}}} 
  & BN-Free           \cite{chen2021bnn}             & 67.31\%    \\
  & ISAL               \cite{liu2021influence}          & 64\%    \\
  & CB AL \cite{bengar2022class}                & 62\% \\
  & IWPS      \cite{aljundi2022identifying}               & 63\% \\
  & CoreGCN              & 69.00\% \\
  & HCD                & 74.28\% \\
  & LCHC                & 74.58\% \\
  & DSAL                & 74.77\% \\
  & \textbf{LCD}        & \textbf{74.86\%} \\
\hline
\multirow{6}{*}{\textbf{ResNet-50 \cite{he2016deep}}} 
  & SG AL   \cite{ledinauskas2020trainingdeepspikingneural} & 58.5\%    \\
  & Proxy AL  \cite{coleman2020selection}    & 61\%    \\
  & HCD                & 78.01\% \\
  & LCHC                & 78.28\% \\
  & DSAL                & 78.37\% \\
  & \textbf{LCD}        & \textbf{78.41\%} \\
\hline
\multirow{6}{*}{\textbf{DenseNet-121 \cite{huang2017densely}}} 
  & BNN-VR            \cite{rakesh2021efficacy}    & 60.00\%    \\
  & Cross Fusion  \cite{dealmeida2021turning}       & 59\%  \\
  & HCD                & 78.01\% \\
  & LCHC                & 78.16\% \\
  & DSAL                & 77.99\% \\
  & \textbf{LCD}        & \textbf{78.49\%} \\
\hline
\multirow{6}{*}{\textbf{Swin \cite{liu2021swin}}} 
  & GradGAT        \cite{dihin2023implementation}      & 46.1\%    \\
  & EswinT    \cite{zhoureplication}  & 44\%    \\
  & HCD                & 64.72\% \\
  & LCHC                & 63.91\% \\
  & DSAL                & 64.77\% \\
  & \textbf{LCD}        & \textbf{65.37\%} \\
\hline
\end{tabular}
\label{table_cifar-100_results}
\end{table}

\begin{table}[h]
\centering
\caption{Performance of the proposed hybrid sampling methods compared against the state-of-the-art active learning methods applied to SVHN. Bold denotes the sampling scheme securing the highest accuracy for the corresponding deep network.}
\begin{tabular}{|l|p{0.3\linewidth}|r|}
\hline
\textbf{Models} & \textbf{Methods} & \textbf{Test Accuracy} \\
\hline
\multirow{6}{*}{\textbf{VGG-16 \cite{simonyan2015very}}} 
  & RAL \cite{sun2019active}                & 94\% \\
  & Core set \cite{sener2018active}         & 95\% \\
  & HCD                & 95.49\% \\
  & LCHC               & 95.44\% \\
  & DSAL               & 95.31\% \\
  & LCD                & \textbf{95.63\%} \\
\hline
\multirow{6}{*}{\textbf{ResNet-18 \cite{he2016deep}}} 
  & RS  \cite{beck2021effective}            & 95.1\% \\
  & HCD                & 95.41\% \\
  & LCHC               & 95.19\% \\
  & DSAL               & 95.45\% \\
  & LCD                & \textbf{95.56\%} \\
\hline
\multirow{5}{*}{\textbf{ResNet-50 \cite{he2016deep}}} 
  & \textbf{DRL} \cite{sun2019active}       & \textbf{97\%} \\
  & HCD                & 95.94\% \\
  & LCHC               & 95.33\% \\
  & DSAL               & 95.61\% \\
  & LCD                & 95.93\% \\
\hline
\multirow{7}{*}{\textbf{ResNet-56 \cite{he2016deep}}} 
  & VI  \cite{gudovskiy2020deep}            & 85.7\% \\
  & HCD                & 95.61\% \\
  & LCHC               & 95.33\% \\
  & DSAL               & 95.61\% \\
  & LCD                & \textbf{94.62\%} \\
\hline
\multirow{7}{*}{\textbf{MobileNet \cite{howard2017mobilenets}}} 
  & MixMatch \cite{song2019combining}       & 76.3\% \\
  & WAAL   \cite{shui2020deep}              & 93.61\% \\
  & HCD                & 94.91\% \\
  & LCHC               & 94.73\% \\
  & DSAL               & 94.04\% \\
  & LCD                & \textbf{94.97\%} \\
\hline
\multirow{6}{*}{\textbf{DenseNet-121 \cite{huang2017densely}}} 
  & CUCB  \cite{spasov2019dynamic}          & 94\% \\
  & HCD                & 96.19\% \\
  & LCHC               & 96.10\% \\
  & DSAL               & 96.09\% \\
  & LCD                & \textbf{96.38\%} \\
\hline
\end{tabular}
\label{table_SVHN_results}
\end{table}

\begin{table}[h]
\centering
\caption{Performance of the proposed hybrid sampling methods compared against the state-of-the-art active learning methods on Tiny ImageNet. The top performing methods are highlighted in bold.}
\label{table_tin_results}
\begin{tabular}{|l|p{0.3\linewidth}|r|}
\hline
\textbf{Models} & \textbf{Methods} & \textbf{Test Accuracy} \\
\hline

\multirow{9}{*}{\textbf{ResNet-18 \cite{he2016deep}}} 
  & Zebra \cite{shih2020zebra} & 61.46\% \\
  & NUIA \cite{jain2025non} & 63.62\% \\
  & STAL \cite{bengar2021reducinglabeleffortselfsupervised} & 52\% \\
  & HCD & 63.44\% \\
  & LCHC & 64.18\% \\
  & DSAL & 64.15\% \\
  & \textbf{LCD} & \textbf{64.87\%} \\
\hline

\multirow{7}{*}{\textbf{ResNet-50 \cite{he2016deep}}} 
  & DQAS \cite{zhao2024dataset} & 64\% \\
  & HCD & 66.40\% \\
  & LCHC & 66.31\% \\
  & DSAL & 66.46\% \\
  & \textbf{LCD} & \textbf{66.57\%} \\
\hline

\end{tabular}
\label{table_tiny_results}
\end{table}

\begin{table}[h]
\centering
\caption{Performance of the proposed hybrid sampling methods compared against the state-of-the-art active learning methods applied to PASCAL VOC 2012. The top performing methods are highlighted in bold.}
\label{table_pascal_results}
\begin{tabular}{|l|p{0.3\linewidth}|r|}
\hline

\textbf{Models} & \textbf{Methods} & \textbf{Test Accuracy} \\
\hline

\multirow{9}{*}{\textbf{VGG-16 \cite{simonyan2015very}}} 
  & GMM \cite{agarwal2020contextual} & 70.19\} \\
  & CDAL-RL \cite{choi2021active} & 72\} \\
  & CDAL-CS \cite{choi2021active} & 71\} \\
  & FSSD \cite{li2017fssd} & 82.7\} \\
  & FFSSD \cite{cao2018feature} & 80.8\} \\
  & HCD & 84.37\% \\
  & LCHC & 84.38\% \\
  & DSAL & 84.37\% \\
  & \textbf{LCD} & \textbf{84.44\%} \\
\hline

\end{tabular}
\end{table}

\section{Ablation Study}
\begin{figure}
\centerline{\includegraphics[width=0.5\textwidth]{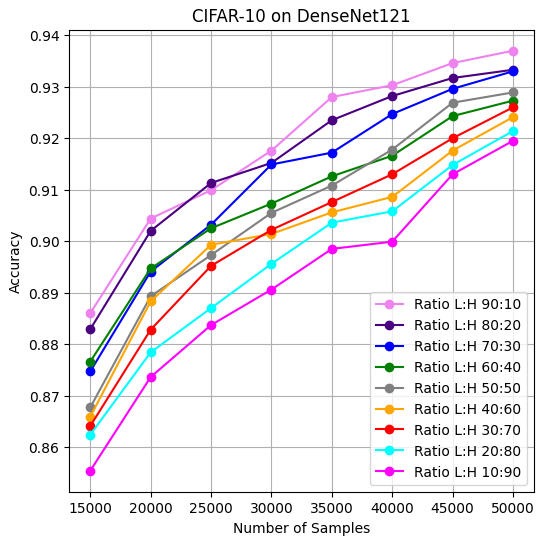}}
\caption{CIFAR-10 test accuracies achieved using DenseNet-121 by selecting the samples upon varying the ratios of hard diverse and easy diverse (DSAL).}
\label{figure2}
\end{figure}

We conduct several ablation studies to observe the performance of the proposed sampling schemes by varying experimental scenarios. The ablation study demonstrates that the proposed hybrid sampling method, DSAL, which selects both diverse least confidence and diverse high confidence samples in a 50:50 ratio, achieves a balanced performance compared to the remaining ratios of DSAL samples. Figure \ref{figure2} shows the performance comparison of DenseNet-121 on CIFAR-10 for different ratios of low confidence and high confidence samples. These findings emphasize the importance of selecting the right ratio of samples in DSAL. The balanced 50:50 ratio effectively leverages the strengths of both uncertainty and certainty sampling strategies, providing the model with informative samples that enhance learning efficiency and robustness. This hybrid method not only improves model performance but also optimizes the labeling budget, making it a valuable strategy for deep learning tasks that require large amounts of labeled data. From Figure \ref{figure2}, it can also be observed that choosing the optimal portion of least and high confidence samples leads to an improvement of 1.23\% compared to selecting 50:50 least and high confidence samples.  

\begin{figure}
\centerline{\includegraphics[width=0.5\textwidth]{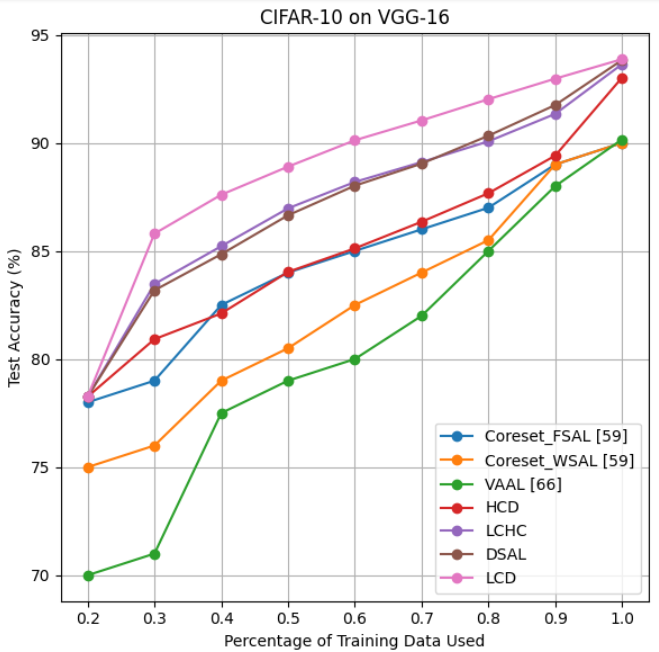}}
\caption{Comparison of different active learning methods on CIFAR-10 using the VGG-16, with the X-axis showing the percentage of training data and the Y-axis representing the test accuracy. The LCD method achieved 86.8\% test accuracy with just 30\% of training samples, which is 7\% more than the core-set method \cite{sener2018active}.}
\label{fig:compare_state_of_the-art}
\end{figure}

We have compared the proposed LCD active learning method against the state-of-the-art methods such as \cite{sener2018active,sinha2019variational} in Figure \ref{fig:compare_state_of_the-art}. From Figure \ref{fig:compare_state_of_the-art}, it is evident that the LCD method achieved 85.5\% test accuracy with just 30\% of the training samples, outperforming all other AL methods. Further, the accuracy of LCD is about 6\% higher than the core-set selection method.

\begin{figure*}[htbp]
  \centering
  \begin{subfigure}{0.48\textwidth} 
    \includegraphics[width=\linewidth, height=4cm]{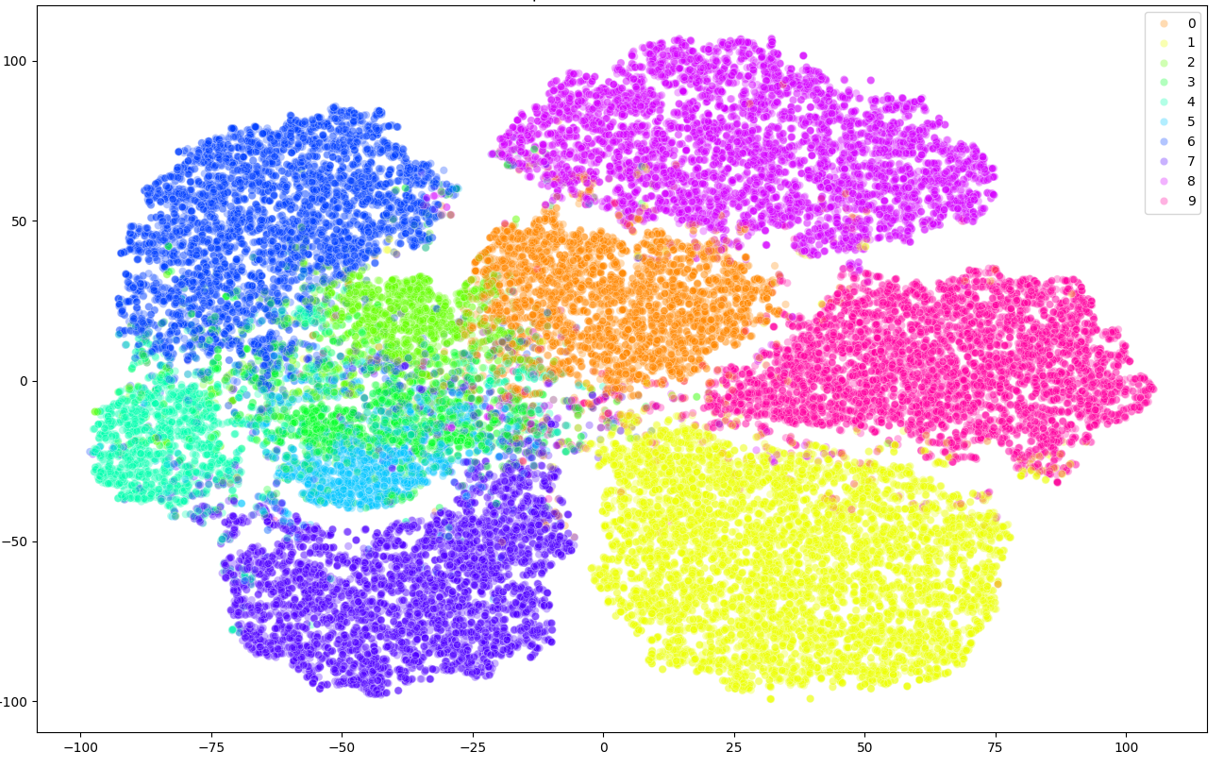} 
    \caption{HCD}
    \label{subfig:y1}
  \end{subfigure}
  \hfill
  \begin{subfigure}{0.48\textwidth} 
    \includegraphics[width=\linewidth, height=4cm]{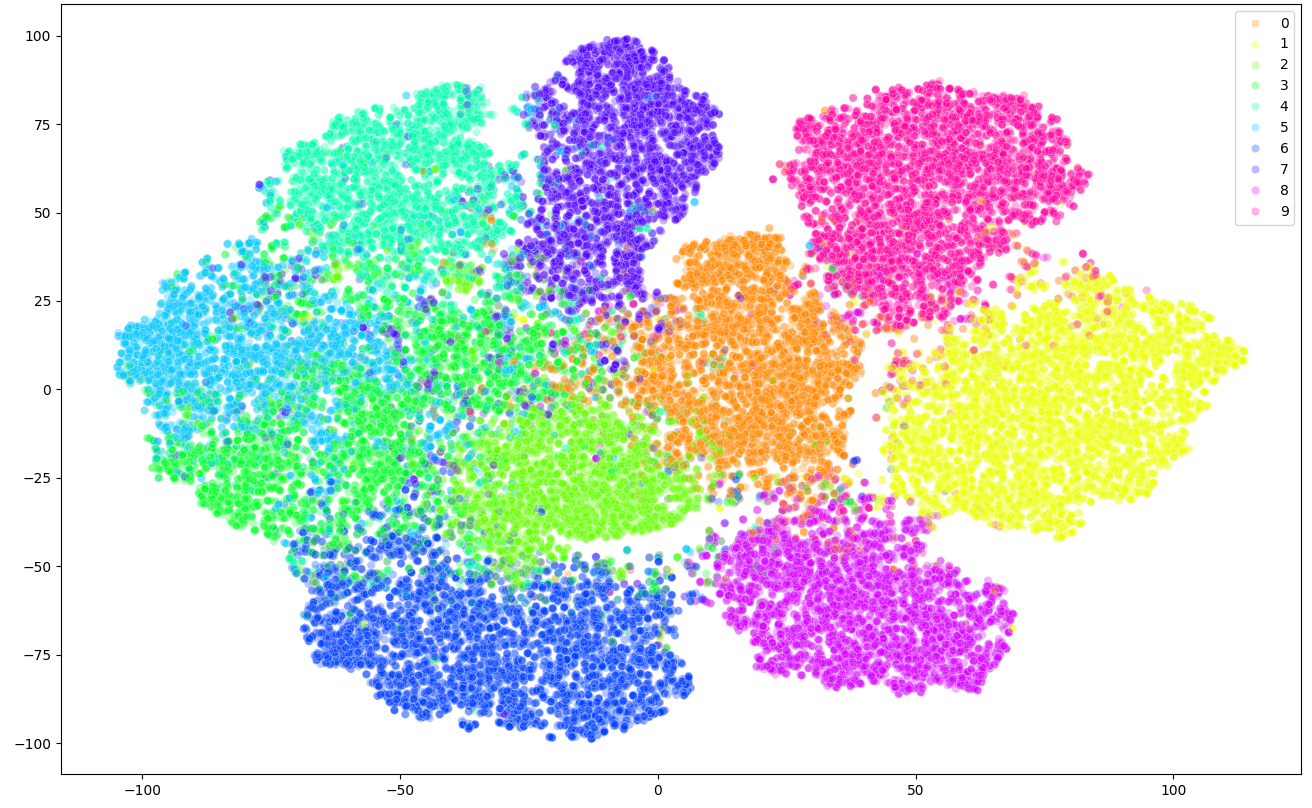} 
    \caption{LCHC}
    \label{subfig:y2}
  \end{subfigure}
  
  \vspace{0.5em}
  
  \begin{subfigure}{0.48\textwidth} 
    \includegraphics[width=\linewidth, height=4cm]{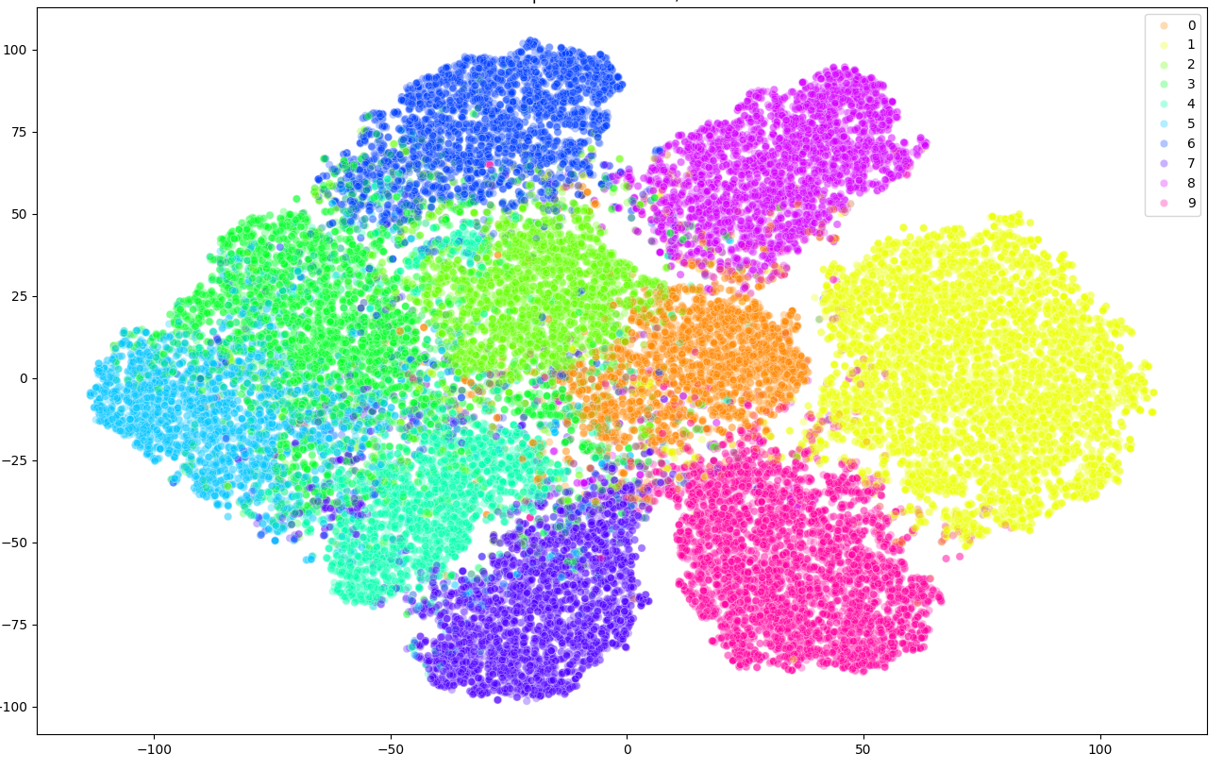} 
    \caption{DSAL}
    \label{subfig:y3}
  \end{subfigure}
  \hfill
  \begin{subfigure}{0.48\textwidth} 
    \includegraphics[width=\linewidth, height=4cm]{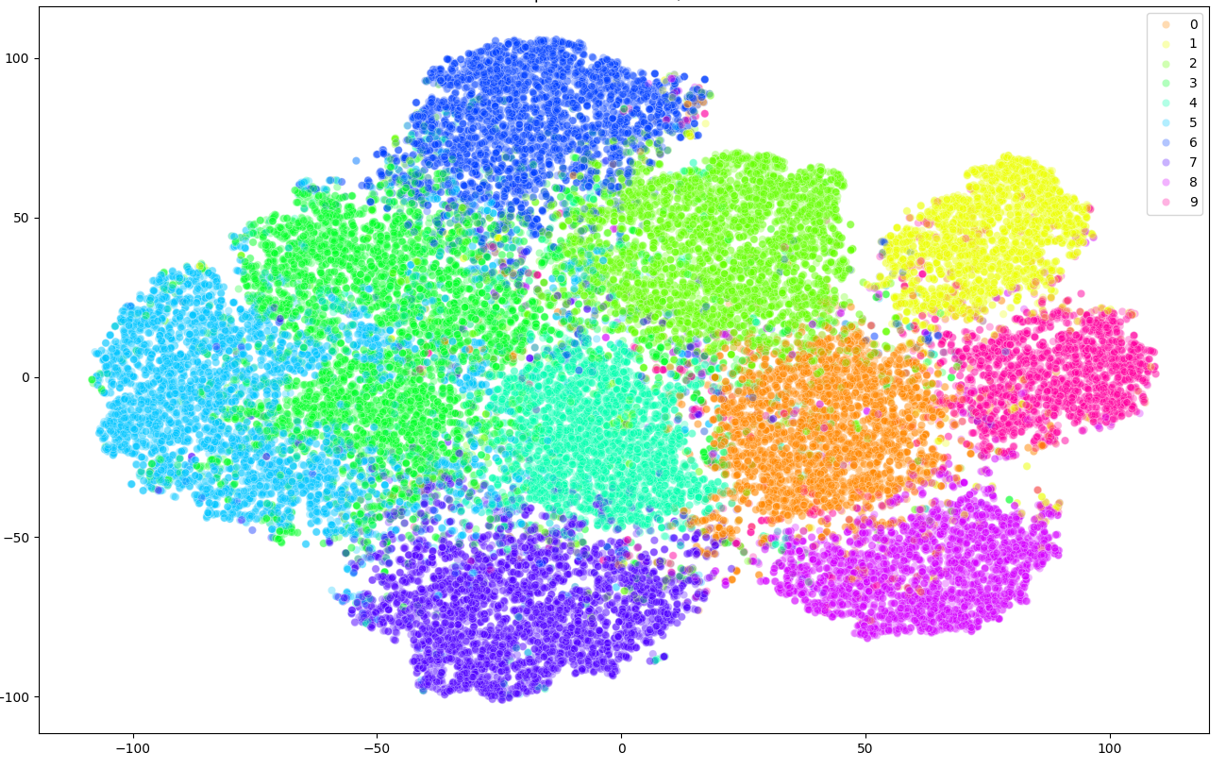} 
    \caption{LCD}
    \label{subfig:y4}
  \end{subfigure}
  \caption{t-SNE plots of the CIFAR-10 training samples in the feature space for different active learning methods. The feature space clearly separates the training instances belonging to each class in increasing order for HCD, LCHC, DSAL, and LCD, respectively. Moreover, the separation of clusters is more prominent for LCD compared to the other methods, which supports the results shown in Tables \ref{table_cifar-10_results}, \ref{table_cifar-100_results}, and \ref{table_SVHN_results}.}
  \label{fig:tsne_plots}
\end{figure*}

Figure \ref{fig:tsne_plots} shows the t-SNE plots, which depict the clustering of samples based on four different sampling methods: HCD, LCHC, DSAL, and LCD. Each method incorporates either low-confidence (LC) or high-confidence (HC) samples, or both, along with or without diverse samples. Analyzing these visualizations provides insights into the performance and distribution characteristics of each method.

\subsection{Observation with HCD}
\textbf{Description:} selects high confidence samples that are diverse as well.

\textbf{Observations:} The clusters are less distinct and more dispersed compared to other methods, with a higher degree of overlap among them.
\textbf{Performance:} This method performs the worst. The model may struggle to generalize well because it heavily relies on high-confidence samples, which may not be as challenging, leading to poor performance on unseen data.

\subsection{Observation with LCHC}
\textbf{Description:} It combines low-confidence and high-confidence samples.

\textbf{Observations:} The t-SNE plot reveals distinct clusters with moderate overlap. While clusters are identifiable, there is overlap at the boundaries.

\textbf{Performance:} This method performs reasonably well, balancing between the extremes of confidence. It ensures a mix of hard and easy samples for the model, allowing for better generalization.

\subsection{Observation with DSAL}
\textbf{Description:} It includes a balanced mix of hard and diverse, and easy and diverse samples.

\textbf{Observations:} The clusters are distinct, but there is a moderate level of overlap, suggesting that the model performs well but has room for improvement.

\textbf{Performance:} This method shows descent performance, benefiting from the balanced inclusion of diverse, low-confidence, and high-confidence samples. However, it does not outperform the LCD method.

\subsection{Observation with LCD}
\textbf{Description:} Chooses low-confidence samples that are diverse.

\textbf{Observations:} The clusters are well-formed and distinct with minimal overlap, indicating effective differentiation between different classes by the model.

\textbf{Performance:} This method consistently outperformed other sampling methods and the state-of-the-art. The inclusion of low-confidence (uncertain) samples introduces challenging scenarios for the model, and when combined with diverse samples, it enhances the model's performance by learning unique features that help to generalize well across varied data.

Incorporating low-confidence samples into an active learning process proves beneficial, according to our study. We train the model with challenging scenarios, thereby enhancing its overall robustness. Conversely, relying heavily on high-confidence samples can limit the model's exposure to diverse, challenging instances and may negatively impact performance. These insights underscore the importance of carefully selecting and balancing sample types to enhance performance.

\section{Conclusion}
\label{Conclusion}
We conduct rigorous experiments with different hybrid sampling schemes for the active learning paradigm. Our study demonstrates that hybrid sampling methods incorporating both uncertainty and diversity can significantly improve the performance of deep learning models in active learning scenarios. The proposed hybrid approach makes sure that all of the unlabeled data space is explored by taking into account both the distance from cluster centers (diversity) and the confidence score of the samples (uncertainty). Furthermore, utilizing higher-dimensional representations of the data enables to capture of rich information and potentially leads to even better performance. This paves the way for future research that explores the effectiveness of hybrid sampling methods in active learning paradigms, in even more complex and high-dimensional applications involving images, videos, and even multi-modal data. To further enhance the practicality of our framework, future work will explore computational optimizations for large-scale data. Efficient strategies such as fast approximate clustering will help make active learning more scalable without sacrificing effectiveness.

{\small
\bibliographystyle{ieee_fullname}
\bibliography{egbib1}
}

\end{document}